# Emulate Randomized Clinical Trials using Heterogeneous Treatment Effect Estimation for Personalized Treatments: Methodology Review and Benchmark

Yaobin Ling M.S., Pulakesh Upadhyaya Ph.D., Luyao Chen M.S., Xiaoqian Jiang Ph.D., Yejin Kim Ph.D.

School of Biomedical Informatics, University of Texas Health Science Center at Houston, Fannin 7000, Houston, Texas

Telephone number: 713 500 3998
Fax number: 713 500 0360
E-mail address: yejin.kim@uth.tmc.edu

## ABSTRACT

Big data and (deep) machine learning have been ambitious tools in digital medicine, but these tools focus mainly on association. Intervention in medicine is about the causal effects. The average treatment effect has long been studied as a measure of causal effect, assuming that all populations have the same effect size. However, no "one-size-fits-all" treatment seems to work in some complex diseases. Treatment effects may vary by patient. Estimating heterogeneous treatment effects (HTE) may have a high impact on developing personalized treatment. Lots of advanced machine learning models for estimating HTE have emerged in recent years, but there has been limited translational research into the real-world healthcare domain. To fill the gap, we reviewed and compared eleven recent HTE estimation methodologies, including meta-learner, representation learning models, and tree-based models. We performed a comprehensive benchmark experiment based on nationwide healthcare claim data with application to Alzheimer's disease drug repurposing. We provided some challenges and opportunities in HTE estimation analysis in the healthcare domain to close the gap between innovative HTE models and deployment to real-world healthcare problems.

**Keywords**

Causal inference, Target trial, Conditional average treatment effect, Drug development, Deep learning, Machine learning

# 1. Introduction

Causal inference discovers a cause of an effect. Although randomized experiments (e.g., randomized clinical trials, A/B test) are a de facto gold standard to identify causation, they are sometimes economically infeasible or unethical if intervention harms subjects [1]. The treatment effect estimation using observational data (e.g., real-world data) is an alternative strategy to emulate the randomized experiments and infer the causation. However, observation inevitably contains bias. A *confounding variable* is a variable that influences exposure to the treatment and outcomes(Fig. 1A). It is one of the major sources of bias that can mislead us to draw a wrong conclusion that the treatment has effects on the outcome when it does not.[2] A statistical approach to reducing such bias in observational data for treatment effect estimation has long been studied in multiple disciplines. For example, a *target trial* framework in epidemiology and biostatistics has been focused on hypothesis testing to infer the *average* treatment effect by adjusting the confounders via matching or weighting [3–12] (Fig. 1Bc, 1Bd)

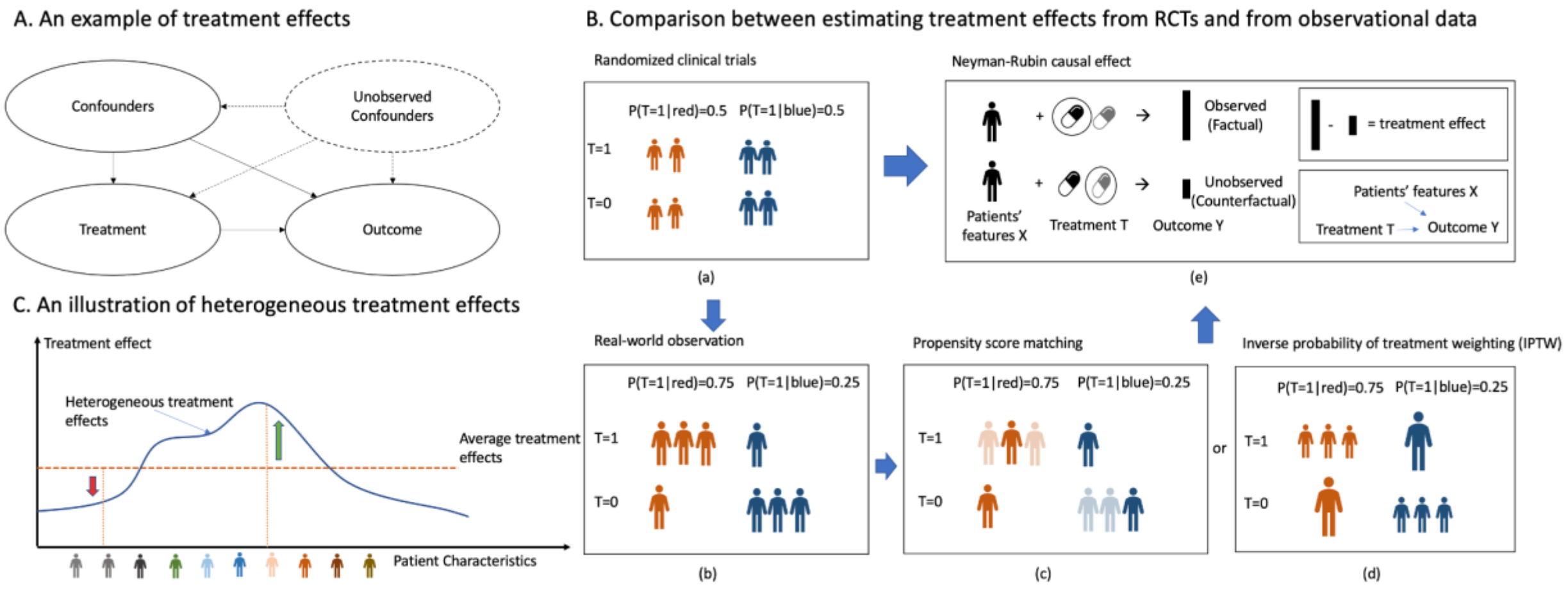

**Figure 1**. Illustrations of treatment effects analysis in the medical science field. A. An example of a causal relationship. B. Estimating causal treatment effects from real-world data under the Neyman-Rubin framework. (a) Subjects in RCTs are randomly assigned to a treatment group and a control group, thus the subjects in both groups have similar characteristics. (b) Subjects in real-world data are not randomly assigned to a treatment group and control group due to disease indication. (c) Matching subjects in each group can reduce bias [13]. (d) Weighting subjects by their propensity for treatment can create a comparable pseudo population [14,15] (Details described in S.1.1). (e) Neyman-Rubin causal effect calculation. C. Heterogeneous treatment effects vs. Average treatment effect. Patients are diverse and treatment effects vary. Estimating the *average* treatment effect (ATE) may oversimplify the heterogeneity of each patient.

However, patients are diverse and treatment effects vary. Decades of drug development in complex diseases have shown that there is no “one-size-fits-all” treatment [16]. Estimating the *average* treatment effect (ATE) may oversimplify the heterogeneity of each patient. The need for personalized treatment is tremendous. Therefore, it is important to estimate treatment effects for each individual or similar subgroups of patients, which is the so-called *heterogeneous treatment effect (HTE)* (Fig. 1C). The HTE estimation has transformative potential in personalized medicine by respecting the disease and patients' heterogeneity.

The HTE estimation is recently gaining attention in econometrics [17] (e.g. uplift modeling), and machine learning (non-parametric HTE) [18], but is rarely investigated in the computational medicine area [19]. To close the gap in this translational effort, we review and compare some recent HTE methodologies and perform benchmark experiments to test the feasibility of the methodologies in emulating clinical trials for personalized treatment development. Our benchmark experiments use nationwide electronic health records with ~60M patients in the US under the target trial protocol. Our scope is within a translational biomedical research perspective, particularly with digital health. This review and benchmark paper adapts notations and naming strategies from various sources including *econml [20]*, *causalml [21]*, Künzel et al. [22], and Bica et.al. [19]. For a theoretical and methodological comparison, see [23–25]. For an econometric perspective, see [26]. For a clinical pharmacology perspective, see [19].

# 2. Preliminaries

### 2.1 Potential outcome framework

In this paper, we investigate models built under Neyman-Rubin's potential outcome framework [27,28]. Suppose that we have $N$ subjects ($i = 1, \cdots, N$) with the feature $X$. For each patient, $T$ denotes treatment assignment; $T = t$ if the subject is in the treatment group *t* with the potential outcome $Y(t)$, and $T = t_0$ if the subject is in the control (placebo) group with the potential outcome $Y(t_0)$. The Neyman-Rubin framework is to estimate the treatment effect of treatment *t* given subject feature *X* (Fig. 1Be) by

$$\tau(x) = E[Y(t) - Y(t_0)|X] \quad (1)$$

The potential outcome framework requires several assumptions:

- *Strong ignorability (or exchangeability)*. We assume no unobserved confounders exist or we observe all the variables $X$ affecting treatment assignment $T$ and outcomes $Y$, i.e., $Y(t) \perp T|X$,[27,29]. Take the language of real-world drug administration as an example, we observe a sufficient set of confounding variables on patient characteristics that determine the outcome and administration of drugs.
- *Positivity.* The probability, $P(T|X)$, of receiving the treatment is not deterministic. i.e., $0 < P(T|X) < 1$ [30]. For example, a patient's feature *X* does not 100% guarantee the onset of a particular specific drug.
- *Stable Unit Treatment Value Assumption*. Each subject's potential outcomes remain the same regardless of what treatment the other subjects receive (no interference between the subjects).[27] For example, a patient taking a drug does not affect other patients' choice of drug.

### 2.2 Counterfactual outcome

A fundamental challenge of the treatment effect estimation is that it is impossible to observe $Y(t)$ and $Y(t_0)$ simultaneously. We call the outcome that the subject has a *factual outcome* and the other hypothetical outcomes in an alternative situation a *counterfactual outcome*. A common approach to address this missing counterfactual outcome is to calculate the average of the two potential outcomes in the treatment group and control group separately after randomization. We estimate the average treatment effect (ATE) of treatment by $E[Y(t)] - E[Y(t_0)]$ if the treatment assignment is randomized. However, in real-world observational data, patients take drugs based on indication, not at random. The patients exposed to the drugs and those not exposed to the drugs are not equivalent. The confounding variable creates a selection bias between the treated and untreated. Several techniques to adjust the confounding variables use propensity scores$e(x) = P(T|X)$, such as matching (Fig. 1Bc), stratification, and inverse weighting (Fig. 1Bd), doubly robust estimations [31], or conditional independence (g-estimation) [32]. More details in Supplementary Material S.1.1.

## 3. Heterogeneous treatment effects

### 3.1 Overview

Heterogeneous treatment effect estimation is to quantify individual or subgroups' treatment effect by accounting for the heterogeneity of patient's conditions to outcome while reducing selection bias. The HTE of treatment *T*=*t* given the patient's condition $X$ is:

$$\tau(x) = E[Y(t) - Y(t_0)|X = x], \qquad (1)$$

which varies based on the subject's features *X*. A general step to estimate the HTE is to first learn the "nuisance" or "context" (likelihood of being exposed to treatment $T$ and expected values of outcomes *Y* at given subject *X*) via arbitrary supervised models (Fig. 2A and Fig. 2B). Then it estimates the HTE by learning a coefficient of a structural equation model or imputing missing counterfactual outcomes in non-parametric methods (Fig. 2C). Many HTE estimation methods have been proposed [18,22,33–37]. Among the models, double machine learning, meta-learners, representation learning, and causal forests are the most popular methods and are flexible to be implemented in many scenarios. We will focus on these three methods, and they were summarized in Table 1.

**Table 1**. Comparison of HTE estimation methods.

| **Models** | **Characteristics** | **Treatment type** | **How is the treatment assignment incorporated?** | **How is the selection bias handled?** |
|---|---|---|---|---|
| Double machine learning [33] | Learn HTE by the coefficient of a partially linear structural equation<br><br>Widely used in the econometric community | Continuous, discrete | T is a function of X to learn. | T is treated as a function of X to learn. |
| Single learner | Learn single outcome prediction model for both the treated and untreated group (E[Y\|X,T])<br><br>The treated and untreated group share the same regression models | Continuous, discrete | T is a variable (together with X) to predict outcome | Not considered |
| Two-learner | Learn multiple outcome prediction models for each treatment group. | Discrete | Separate outcome prediction model based on T | Not considered |
| X-learner [22] | T learner with additional treatment effect estimation regression models for each of the treated and untreated groups.<br><br>Handle data size imbalance in the treated and untreated group | Discrete | Separate outcome prediction model based on T | Weighted average of each treatment effect estimation based on propensity scores |
| BNN [34] | Minimize the distribution distance between the transformed representation of the treated and untreated | Binary | Not considered | Covariate shift |
| TARNET [18] | BNN with multi-task heads for each treatment group, without covariate shift | Binary | Separate potential outcome prediction layers on each T | Not considered |
| CFR [38] | BNN with multi-task heads for each treatment group | Binary | Separate potential outcome prediction layers on each T | Covariate shift |
| DR-CFR [35] | Separate representations for propensity score and outcomes prediction | Binary | Separate potential outcome prediction layers on each T | Covariate shift |
| SCIGAN [36] | Estimate counterfactual outcomes via generative adversarial networks. | Continuous, discrete | T is a variable to predict dose-dependent outcome | Covariate shift |
| Dragonnet [37] | Predict Y using X that are predictive of T (Sufficiency of Propensity Scores). Regularize the outcome prediction to satisfy augmented IPTW estimator | Discrete | Separate potential outcome prediction layers on each T | Learn neural network to predict outcome satisfying augmented IPTW estimator |
| Double sample Causal Tree [39] | Use an ‘honest’ method to grow a tree. The splitting criterion is to maximize the variance of estimated treatment effects among samples in the training subset. | Discrete | Not considered; only for randomized data, in which treatment assignment is fully randomized | Honest splitting criterion and cross-validation |
| Causal Forest[40] | Using multiple causal trees to get predictions of treatment effects (ensembling). Point estimates from the model are asymptotically normal and unbiased, and so allow for confident intervals to be calculated. | Discrete | Not considered; only for randomized data, in which treatment assignment is fully randomized | Honest splitting criterion and cross-validation |

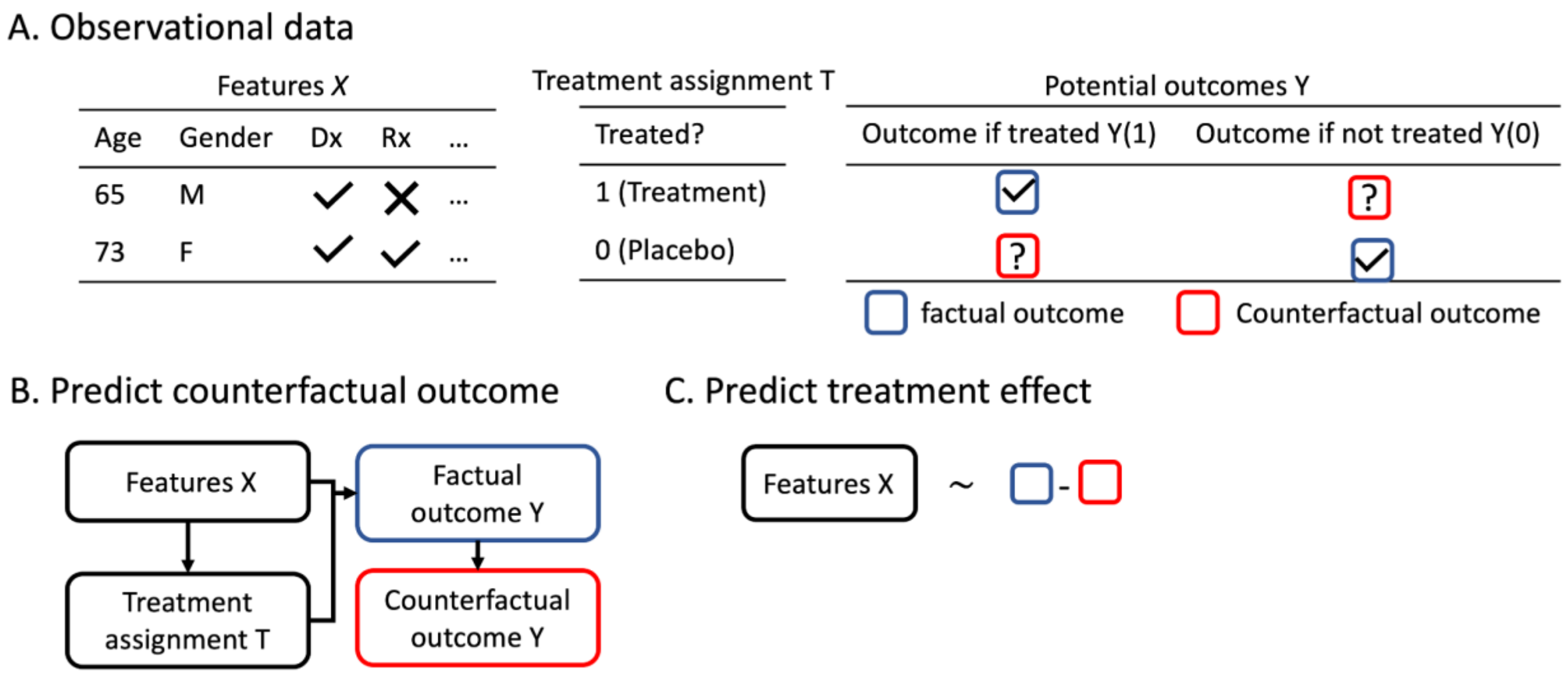

**Figure 2**. A toy example of heterogeneous treatment effects with high-dimensional features. Based on the subject's feature $X$, the subject can be exposed to treatment $T$ or not (i.e.,$E[T|X]$) and has different levels of outcomes (i.e.,$E[Y|X]$). The treatment effects also vary based on these different contexts or nuisances stemming from the subject's features.

### 3.2 Double machine learning

Double machine learning shows us how the parametric causal inference evolves to semi-parametric causal inference with machine learning. A challenge that classical causal inference faces are the high-dimensional and complex nuisance parameter around the causal estimates. The complex nuisance parameters are often difficult to be estimated with a classical semi-parametric framework (e.g., 10 binary features mean $2^{10}$ combinations of features to test whether the treatment effects are distinctive from the others). To address the high dimensionality of features, Chernozhukov proposes to replace semi-parametric nuisance with arbitrary supervised machine learning models, which is the so-called Double machine learning [33]. The Double machine learning architecture is composed of two arbitrary supervised machine learning models which are used to estimate the conditional probability of taking the treatment and the conditional expectation of the outcome respectively. Then one can estimate the HTE by fitting a regression model (Details in Supplementary material S.1.2). The architecture was later generalized to R-learner and was proved to have an asymptotic error rate.

### 3.3 Meta-learners

Meta-learners are one of the nonparametric HTE estimation methods that treat the HTE estimation for discrete treatment as missing counterfactual outcome imputation. They decompose the HTE estimation into several sub-regression problems that any arbitrary supervised machine learning model can be utilized. We follow the naming strategy introduced by Künzel et.al.[22]

- Single-learner (or S-learner), the most basic method in meta-learner, treats the treatment assignment variable $T$ as just another feature (in addition to $X$) and builds a "single" supervised model to estimate the outcomes $Y$ (Fig. 3Aa).
- Two-learner (T-learner) fits separate regression models for treatment and control groups respectively (Fig. 3Ab). The advantage of T-learner over S-learner is that T-learner performs relatively better when there is not much similarity between the outcome given treatment and the outcome given the control [22].

- X-learner is a variant of the T-learner with extra steps to separate the HTE functions (Fig. 3Ac).[22] It is named after “X”-like shaped use of training data for counterfactual outcome estimation. The main advantage of X-learner over T-learner is that X-learner separates the treatment effect regression of the treatment and placebo group so that the HTE estimators can capture information about their differences. This strategy is beneficial when the number of subjects treated and untreated is not balanced.

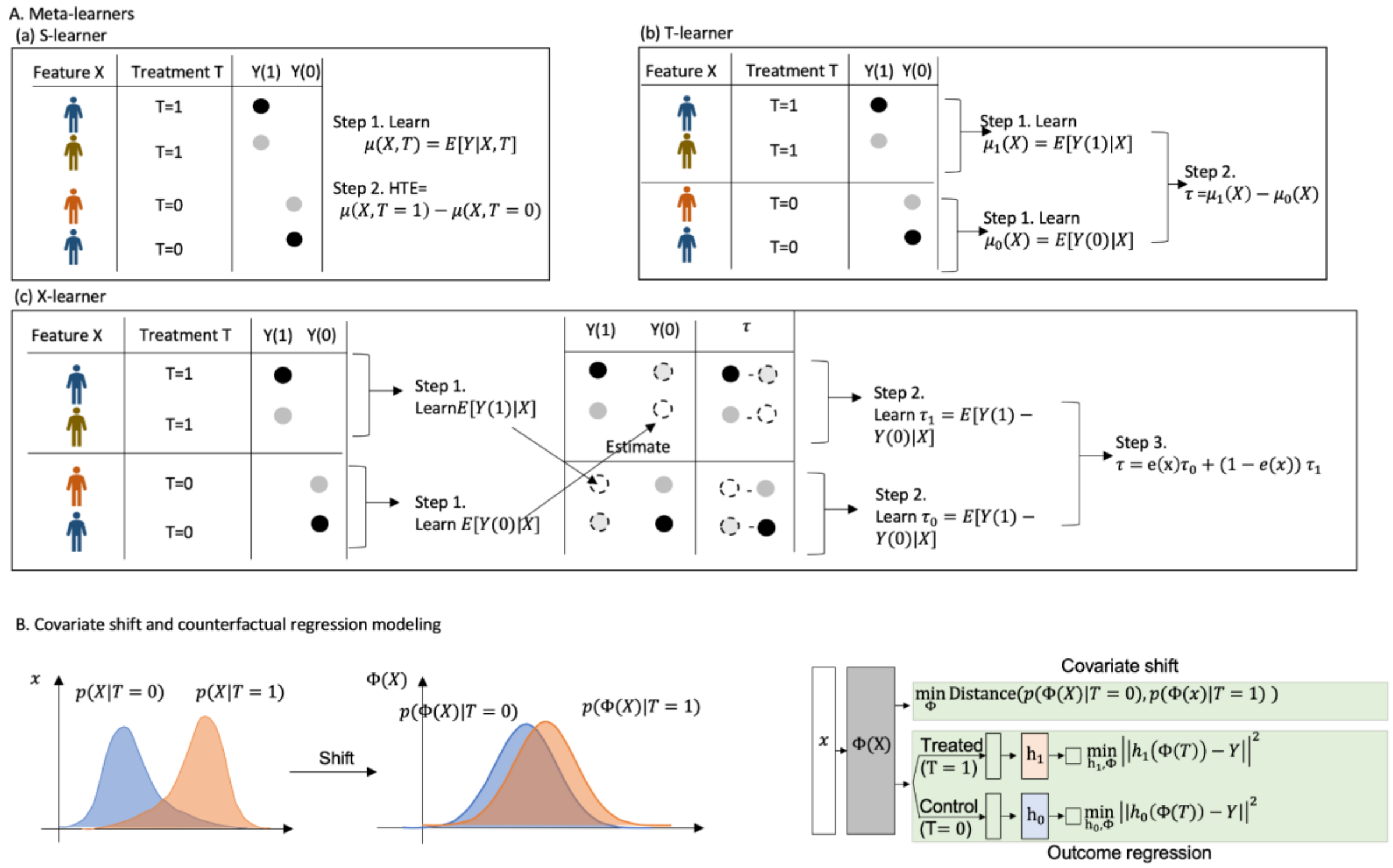

**Figure 3**. Illustrations of different architectures of HTE models (A) A toy example to compare meta-learners in Künzel et.al.[22] Metalearners decompose the HTE estimation into several sub-regression problems that any arbitrary supervised machine learning model can be utilized. (B) Illustration of covariate shift problem and counterfactual regression network [18,38]. Balanced representation learning via covariate shift using neural networks [18,38] with the treatment-invariant representation of a patient’s feature X.

Despite its flexibility and simple intuition, the meta-learner has several limitations. The meta-learners are non-parametric models with full flexibility to select any arbitrary supervised machine learning model, making it difficult to obtain a valid confidence interval. Also, the meta-learners are only available with discrete treatment. With multiple discrete treatments, they require a model for each treatment, posing extra computations. For more details, see [22] and some open-source implementations, *causalml [41]*, and *econml* [42].

### 3.4 Representation learning

Representation learning based on deep neural networks has also been actively used in nonparametric HTE estimation. Recent works have focused on learning a covariate shift function by which feature representation of the treated and untreated follows a similar distribution (Fig. 3B) and studied the trade-off between balance and predictive power of such algorithms. BNN [38], CFRNet [38], and TARNet [18] propose a family of algorithms to predict HTE using the balanced representations learned from observational data (Fig. 3B), with the last two providing a bound for the HTE estimation error. Their theoretical approaches are also applied to many synthetic and real-world datasets [38].

Specifically, the balanced learning approach finds a representation $\phi: X \to R$ and treatment-specific head $h_1$and $h_0$, that will minimize the evaluation measure PEHE:

$$PEHE = \int (\hat{\tau}(x) - \tau(x))^2 p(x) dx \quad (2)$$

where $\hat{\tau}(x) = h_1(\phi(x)) - h_0(\phi(x))$. It utilizes a loss function that is lower bounded by PEHE to train the model; the loss function consists of the sum of the expected factual treated and control losses for outcome regression and distance of $\phi(x)$ given $t = 0$ and $t = 1$ for covariate shift (Fig. 3B).

### 3.5 Tree-based methods

Another class of models is a tree-based model [39,40,43]. Tree-based models, including causal trees and causal forests, are nonparametric models that use recursively splitting criteria to find subgroups in which the sub-samples can be viewed as from randomized experiments and the divergence between outcomes of treatment and control groups is maximized. One characteristic of tree-based models is that they keep good asymptotic properties, and thus allow users to conduct solid statistical inference about the point estimators from these models. In addition, tree-based models can provide generated rules for further interpretation and external validation. [44]

Double sample causal tree, one of the most representative tree methods, uses an 'honest' estimation method to grow a tree. It splits the training random sub-samples into two parts, one is used for predicting outcomes, and the other part is used to find the split for the node. The 'honest' mean that with sample $i$, one can use the response $Y_i$ for estimating treatment effects within the leaf or for finding the split but cannot use it for both. Propensity tree is another method that incorporates the estimation of propensity scores in the model to adjust for confounders in observational datasets [40],

Causal forest ensembles many causal trees and estimates treatment effects by averaging predictions of the ensembled trees. Considering randomness in a single causal tree, one can never know if it is the 'best' tree. By aggregating results from many trees, the causal forest thus can provide more robust estimators and smooth decision rules.

### 3.6 Methodology comparison

The HTE estimation methods we discussed have their strengths and weaknesses. Double machine learning utilizes structural equations to model the discrete or continuous treatment with confidence intervals. Meta-learners allow users to use any base learner to fit outcome prediction and treatment effect regression, so researchers have the autonomy to choose supervised models that best work for their data. The HTE estimation using representation learning is an end-to-end approach that all tasks (reducing selection bias, predicting potential outcomes, and calculating treatment effect) are seamlessly connected in one neural network framework. Thanks to the neural network's flexibility as a function approximator, various modeling hypotheses (e.g., multi-head for respective treatments [18], disentanglements of representation [35].) can be incorporated and tested.

Let us compare the methods based on three criteria: treatment type, treatment assignment, and selection bias (Table 1). For the treatment type, T-learner and X-learner handle binary treatment, but it's straightforward to extend it to multiple discrete treatments. In contrast, BNN, CFRNet, and TARNet assume minimizing the distance between two treatment groups, thus requiring more computational challenge when extending the binary treatment to multiple discrete or continuous treatments. Instead, Double machine learning and S-learner can incorporate continuous treatment. For the treatment assignment variable, most methods for discrete treatments adopt separate outcome prediction (i.e., either neural network layers/heads or an independent prediction model) to handle the different potential outcomes. This setting is advantageous when the size of treatment groups is not balanced. For selection bias handling, covariate shift is the main approach in representation learning [18]; covariate selection by propensity scores was also investigated. Weighting by propensity scores was also widely used [22].

### 3.7 Evaluation metric

Directly evaluating the accuracy of the estimated HTE is challenging because the ground-truth treatment effect is never observed in data; randomized experiments are the only method to obtain the ground truth. Researchers have used several indirect measurements: robustness and estimated goodness-of-fit.

- Robustness: To evaluate whether the model's estimation is robust to different data, one can train the same model on training data and test data respectively, and then calculated the "estimated" root mean squared error (ERMSE) of the estimated treatment effects from training data and test data. High robustness (or low ERMSE) means the model consistently generates a similar HTE estimation regardless of the input data, which supports the validity of the estimation.
- Estimated goodness-of-fit**:** To evaluate how accurately the models can predict HTE, the precision of estimating heterogeneous effects (PEHE) is a direct metric for this goal [45]. The PEHE is defined as the difference between the true HTE and the estimated HTE (Eq. 2). As the true HTE is never observed, Alaa and Schaar proposed the influence function-PEHE (IF-PEHE) that approximates the true PEHE by "derivatives" of the PEHE function, not directly relying on the unobserved counterfactual outcomes [46]. The approximation is composed of two parts: a plugin estimate and an influence function used to compensate for bias. Using a well-designed plugin estimate makes it easier to train and gives a partial guess on the true PEHE; the remaining bias from the plugin estimate is approximated by the influence function, which is analogous to the derivatives of a function in standard calculus. See details in the paper [46] or Supplementary Material S.1.4.

# 4. Benchmark experiments

We tested the feasibility of the HTE estimation in identifying personalized drug effectiveness. We focused on a task to emulate a randomized clinical trial (NCT03991988) [47] that tests the treatment effect of Montelukast on treating AD. Our emulation is based on a target trial framework[11] that mimics the actual trial's eligibility criteria, treatment strategy, and observation period from nationwide large-scale claim data. From the emulated randomized clinical trial, we aim to estimate the HTE of Montelukast (as well as other anti-asthma drugs) in reducing AD risk and ultimately evaluate the feasibility of the HTE estimation as a new tool for identifying personalized drug effectiveness.

### 4.1 Background in AD and Montelukast

AD is a progressive neurologic disorder with the death of brain cells. Chronic inflammation is known to be linked to AD [48,49]. Montelukast is an anti-inflammatory drug for chronic inflammation, such as asthma.  Some researchers speculate that Montelukast might reduce AD risk, not only by reducing chronic peripheral inflammation but also by targeting the leukotriene signaling pathway, which mediates various aspects of AD pathologies [49]. An ongoing Phase II trial is investigating the treatment effect of Montelukast on AD [47]. However, the average cost of such trials is ~13 million dollars in the U.S. [1], and with 99% failure rates in AD treatment trials [49]. Emulating the RCTs via treatment effect estimation before the actual trials may offer a cost-effective and safe tool to reduce the failure rates. However, there are several confounding variables in observational data. If Montelukast and AD turn out to have a significant correlation or association, the high association may or may not be due to the direct preventive effects of Montelukast on AD. It may be due to the anti-inflammatory effects of Montelukast on reducing chronic inflammation. In addition, AD has various pathologies developing clinical symptoms (such as via neuroinflammation, and protein misfolding). The treatment effects of Montelukast on AD vary by the pathology one patient has.

### 4.2 Experimental Setting

We defined eligibility criteria, standard-of-care arm, follow-up period, and outcome based on NCT03991988 [47] (Table 2, Supplementary Material S.2.1).

**Table 2.** Comparison of our experiment setting and the actual trial NCT03991988.[50]

| | Randomized clinical trial on Montelukast (NCT03991988) | Our target trial design |
|---|---|---|
| Aim | Assess the effects of Montelukast initiation on amyloid and tau accumulation and cognitive symptoms of prodromal AD. | Assess the effect of one type of anti-asthma drug on AD onset prevention from routine clinical care setting in real-world data (2007-2020) |
| Eligibility criteria | - Age > 50<br>- Mild cognitive impairments<br><br>**Diagnosis affecting AD:**<br>- No depression, schizophrenia, Parkinson's disease, multiple sclerosis when the trial starts<br>- No history of stroke in the past 3 years.<br>- No history of increased intracranial pressure<br><br>**Indication of Montelukast:**<br>- No comorbidity on bronchial asthma or exercise-induced bronchospasm<br><br>**Adverse effect of Montelukast:**<br>- No liver disease, renal disease due to safety<br>(3 inclusion criteria, 15 exclusion criteria) | - Born before 1942<br>(Age at 2020 > average ADRD onset age) *<br>- No history of AD onset yet at follow-up starts<br><br>**Diagnosis affecting AD**<br>- No depression, schizophrenia, Parkinson's disease, multiple sclerosis when observation starts<br>- No history of stroke in the past 3 years.<br>- No history of increased intracranial pressure |
| Treatment strategies | **Treatment arm:**<br>- Continuous therapy of an anti-asthma drug (Montelukast);<br>- No other drugs within the same category (Leukotriene receptor antagonists)<br><br>**Standard-of-care arm:**<br>- No other drugs within the study drug class | **Treatment arm:**<br>- Continuous therapy of one class of anti-asthma drug<br>- No other drugs within the same category<br><br>**Standard-of-care arm:**<br>- No other drugs within the study drug class<br>- Take other active anti-asthma drugs |
| Assignment procedure | Random assignment | Reduce selection bias computationally (Propensity score, covariate shift in balanced learning) |
| Follow-up period | 2019.09.25 – 2022.10 | Follow-up starts at first records meeting eligibility criteria<br>Follow-up ends at ADRD onset or last observation, whichever occurs first |
| Outcome | -Adverse effect (gastrointestinal symptoms, anaphylaxis, elevated liver enzymes)<br>- CSF Amyloid and tau change<br>- CDR-SB cognitive scores | AD and related dementia onset (PheWas dx codes and medication codes) |

Montelukast is a cysteinyl leukotriene type 1 (CycsLT-1) receptor antagonist to treat asthma and allergy symptoms. Animal model studies show Montelukast's efficacy in reducing amyloid-beta toxicity and neuroinflammation [51,52]. A retrospective study reports an association between Montelukast and reduced uses of dementia medicine in older adults, compared to other anti-asthma drugs [53]. Two Phase 2 clinical trials (NCT03402503, NCT03991988) are ongoing to test the effectiveness of Montelukast on AD's neuropsychological progression.

To emulate clinical trials, we used real-world patient drug claim data. Claim data capture routine clinical care, although there are ongoing debates about whether such real-world healthcare administrative data (including electronic health records and claim data) are suitable data sources to infer treatment effects [54]. We used the Optum Clinformatics® Data Mart subscribed by UTHealth. It comprises administrative health claims from Jan 2007 to June 2020 for commercial and Medicare Advantage health plan members. These administrative claims are submitted for payment by providers and pharmacies and are verified, adjudicated, adjusted, and de-identified before inclusion in Clinformatics ® Data Mart. It contains 6.5 billion claims with 7.6 billion diagnosis codes (in ICD 9/10), 2.7 billion medication codes, and 2.5 billion lab results.

We define the study treatment as leukotriene receptor antagonist (Montelukast, Zileuton, Zafirlukast, and Pranlukast) and standard-of-care treatment as remaining active anti-asthma drugs classes (Beta-2 adrenergic receptor agonist, Anticholinergic drug, Xanthine, and Corticosteroid). See Supplementary Table S1 for specific drug names and their DrugBank ID.

We selected cohorts of older adults meeting the eligibility criteria in NCT03991988 [47]. We included 2,740 patients taking the study treatment and 8,545 patients taking standard-of-care treatments, so the final cohort size was 11,285. We split the subjects into training, test, and validation datasets randomly by 6:2:2 (6,771 for training, 2,257 for validation, and 2,257 for test data).

Potential confounding variables that both affect the exposure to the study treatment and ADRD onset include age when follow-up starts, sex, race, and comorbidities. For comorbidities, we converted ICD9 or ICD10 diagnosis codes into PheWas codes to increase the clinical relevance of the billing codes.[55] PheWas ICD code is a hierarchical grouping of ICD codes based on statistical co-occurrence, code frequency, and human review. A total of 242 PheWas diagnosis codes were included in the dataset. We used the onset of AD as an implicit measure of the increased level of AD risk as the outcome. AD onset was detected as having either an ADRD diagnosis code or medication.

Based on Eq. (1), the HTE is defined as the difference between the estimated AD onset when the patients are intervened to take the study treatment (leukotriene receptor antagonist) and when the patients are intervened to take the control (standard-of-care treatments). A negative HTE value means a reduced AD onset risk due to the study treatment. To evaluate the performance of our HTE estimation model, we measured the robustness (ERMSE) and the estimated goodness-of-fit (IF-PEHE). The detailed plug-in function setting for IF-PEHE is in Supplementary Material S.2.3.

Some HTE estimation methods require a propensity score, the likelihood of a patient to take the study treatment, to weight their estimates and reduce selection bias. We chose the random forest as a propensity score estimation model (Supplementary Material S.2.4). As to estimate propensity score, we used all the features available (the 242 diagnosis codes and three demographics). We checked whether the treatment assignment was at random or affected by other features by calculating the prediction accuracy of the propensity score model (Supplementary Material S.2.4, Table S2).

Meta-learners require several prediction tasks: i) potential outcome classification models for treatment and control groups, respectively, and ii) treatment effect regression (if X-learner and R-learner). We compared logistic regression, random forest, and XGBoost as a choice for each prediction task.

### 4.3 Results

The benchmark experiments show that most HTE estimation models with low variance (S-learners, DML, DragonNet) had better robustness (low ERMSE) and goodness-of-fit (low IF-PEHE). Particularly, S-learner with the Random Forest as the base learner gave the best HTE estimation in both metrics (Table 3). We further discuss the possible reason of the high performance of the simple model in the Discussion section. The representation learning methods had a low ERMSE and high IF-PEHE generally. Although the representation learning methods were not competitive with the low variance models, the DragonNet performs better in both metrics among representation learning methods, because our sparse and noisy input data can benefit from the sufficiency of the propensity score for causal adjustments. Since it uses only the information relevant to the treatment, it gives better estimates when many covariates influence outcomes but have no effect on the treatment.

**Table 3.** Comparison of various HTE estimation models in a target trial for NCT03991988 using nationwide claim data.

| | | Setting | Robustness (ERMSE) | Goodness of fit (IF-PEHE) |
|---|---|---|---|---|
| Meta learners | S learner | Logistic Regression | 0.0225 | 22.12 |
| | | Random Forest | 0.0029 | 9.82 |
| | | XGBoost Classifier | 0.0380 | 19.01 |
| | T learner | Logistic Regression | 0.2933 | 155.63 |
| | | Random Forest | 0.0737 | 21.66 |
| | | XGBoosting | 0.2926 | 57.77 |
| | X learner | Outcome learner: Logistic Regression<br>Effect learner: ElasticNet | 0.0705 | 35.78 |
| | | Outcome learner: Random Forest Classifier<br>Effect learner: Random Forest Regressor | 0.1147 | 22.46 |
| | | Outcome learner: XGBoost Classifier<br>Effect learner: XGBoost Regressor | 0.1980 | 79.33 |
| | R learner | Outcome learner: Logistic regression<br>Effect learner: ElasticNet (equivalent to DML) | 0.0110 | 22.61 |
| | | Outcome learner: XGBoost Classifier<br>Effect learner: XGBoost Regressor | 0.5873 | 385.34 |
| Balanced representation learning | CFRNet | Representation Learner: Deep Neural Network<br>Hypothesis Learner: Deep Neural Network | 0.0547 | 102.60 |
| | DRLearner | Representation Learner: Three neural networks, two regression networks for each treatment arm<br><br>Confounder Learner: two logistic networks to model logging policy, design weights for confounder impact | 0.0597 | 152.99 |
| Representation learning (Others) | DragonNet | Outcome learner: Neural Network Regressor which learns outcome as well as propensity scores.<br>Effect learner: Neural Network Regressor | 0.0401 | 53.57 |
| Tree-based method | Generalized Random Forest | Single tree estimator: double sample causal tree | 0.0060 | 18.90 |

**4.4 Important features in the best model**

Using the best model (S-learner with Random Forest), we investigated patients' features that contribute to the high/low HTE.  We calculated feature importance scores in the HTE estimation using the Shapley scores (Fig. 4).[56] The Shapley score measures the average marginal contribution of a feature by excluding the feature over a varying subset of other features. As a result, age at baseline was listed as the second most important factor; it is shown that older patients benefit less from leukotriene receptor antagonists in reducing the risk of AD. On the other hand, leukotriene receptor antagonists have a beneficial treatment effect on patients with disorders of bone and cartilage (PheWas dx: 733) or gout and other crystal arthropathies (PheWas dx: 274). This finding implies that patients with chronic inflammatory bone and joint disorders are more likely to have beneficial treatment effects from leukotriene receptor antagonists to reduce AD risk. Indeed, there is accumulating evidence suggesting significant interplay between peripheral immune activity (e.g., chronic inflammatory bone and joint disorders), blood-brain barrier permeability, microglial activation/proliferation, and AD-related neuroinflammation [57]. In an *in vivo* study, Montelukast (one type of leukotriene receptor antagonist) inhibits inflammation-induced osteoclastogenesis in the calvarial model [58]. This finding requires in-depth biological investigation.

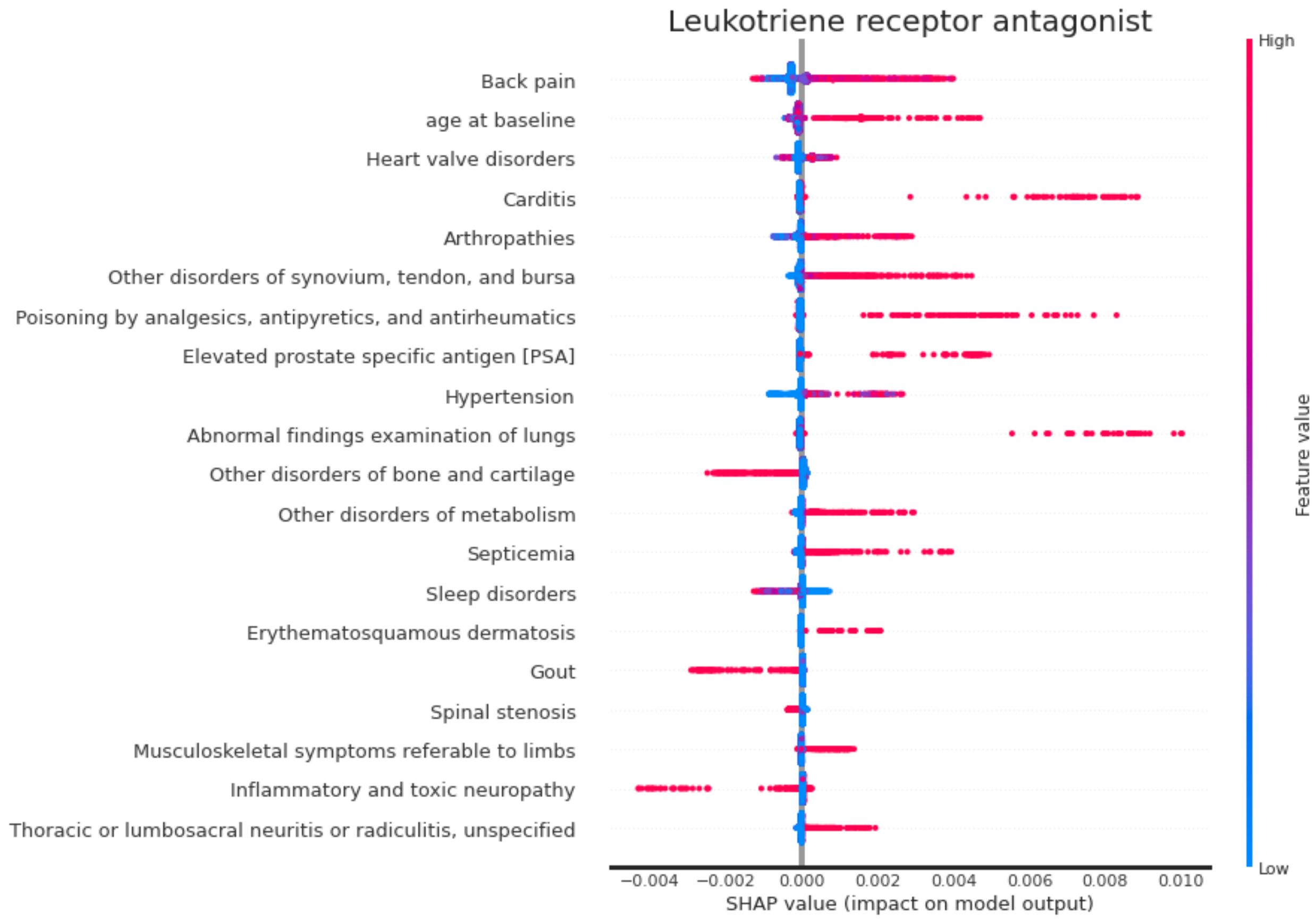

**Figure 4.** Feature importance values of the best-performed model (S-learner with Random Forest). A feature with a negative SHAP value means the feature can lead to negative HTE values, implying a reduced AD onset risk due to the study treatment. Features are sorted in decreasing order by importance. Blue points = low feature value; Red points = high feature values.

# 5. Discussion on the HTE estimation methods

We have reviewed and compared the recent HTE estimation methods to show their feasibility for translational research in biomedicine and drug development. Our extensive benchmark experiments compared several state-of-the-art and popular HTE estimation methodologies such as meta-learners, causal trees, and counterfactual representation learning. As a result, we found that the simple S-learner with Random Forest estimates the HTE most stably with sparse and noisy drug claim data. Possible reasons that the simple model performs best on our sparse claim data are three-fold: i) *shared function in treatment and control groups:* The treatment and control groups might have similar mechanisms between the features and outcome regardless of being treated with the study treatment or not. Incorporating treatment assignment variables as one of the features and building one shared model for the two groups can help models learn the shared mechanisms in the two groups, particularly when the features are sparse, and data are not rich. ii) *Biologically zero treatment effect*: If the study treatment has no treatment effect biologically in nature, S-learner is known to be more accurate in estimating the zero value of HTE [22]. Indeed, there has been no clear scientific evidence that leukotriene receptor antagonists prevent AD. iii) *Avoid overfitting:* S-learner has a small number of model parameters (low variance in predicted values). Possibly S-learner is more likely to avoid overfitting than other meta-learners with a larger number of model parameters (e.g., the X-learner requires three prediction tasks). Our empirical finding is corroborated by prior theoretical investigations on inductive bias in HTE estimation[59]. Based on this finding, we suggest that the future HTE estimation model should consider the shared structure between the arms.

In addition, we found several limitations and challenges in data and problem formulation when applying them to real-world healthcare data and problems. The methods we compared primarily focus on how to achieve a more robust and accurate estimation of HTE through different model architectures, assuming that an ideal form of data is given. However, real-world data are incomplete in observation and lack unbiased control to compare. Here we elaborate on the challenges we identified in detail.

**Unobserved confounding variables.** We assumed that our data contains a sufficient set of confounding variables that determines the outcome (e.g., AD onset) and treatment (e.g., the onset of study treatment). Caution is needed as the assumptions can be violated with real-world drug administrative observational data. Some important confounding variables, such as socioeconomic status, are not included in healthcare administrative data because these data are collected for billing purposes and not for scholarly purposes.

**Variables in healthcare administrative data**

The HTE estimation requires a larger set of variables to capture patients' various conditions that might affect treatment effects at varying levels. Healthcare administrative data (e.g. EHRs, claim data) is a useful data source to contain the various comorbid conditions and co-medication patterns. Two data issues arise, sparsity and multimodality. Healthcare administrative data are sparse, noisy, and missing, not at random. Although we mitigated the sparsity by mapping all the diagnosis codes to PheWas codes[60] to derive compact and dense variables and increase the clinical relevance of the billing-purpose codes, the data was not rich enough, particularly, to train data-hungry deep learning models (e.g., CFRNet, DRLearner, Dragonnet). Previous work on deep learning and healthcare administrative data [61] show that the medical event prediction accuracy of deep learning models is marginal to that of traditional feature-based baselines. This data sparsity challenge might be a barrier to applying the representation learning methods to healthcare administrative data and general machine learning tasks. To mitigate the sparsity of healthcare administrative data, one can consider feature selection. Extra consideration is needed when applying feature selection to some HTE models as they consist of multiple prediction sub-tasks of supervised learning compared to general supervised learning tasks. Several feature selection methods for the HTE estimation were proposed [62]. We tested these feature selections on the best-performing model, S-learner with random forest regression, to see if the feature selection decreased the HTE performance measures, which turns out to be not helpful (Supplementary S.2.5).

Another challenge is data multimodality. Both medication order and diagnosis codes capture a patient's comorbid conditions from different perspectives. One focus of the HTE estimation is to reduce selection bias by non-randomized treatment assignments and incorporating comprehensive multimodal variables. It is a study design choice to determine which modality to include or how to incorporate both modalities that follow different distributions. Healthcare experts' knowledge would also be very critical to selecting the most important variables to predict propensity to treatment.

**Define standard-of-care treatments**

Estimating treatment effect using observational data requires us to define a virtual placebo (or standard-of-care treatments) from which the study treatment is compared. The target trial framework compares the choice of standard-of-care treatment: active drugs vs no active drugs [9,11,12]. Defining "good" standard-of-care treatments is critical that are distinctive enough to compare with the study treatment, not co-administered with the study treatment (avoid patients taking both the study treatment and standard-of-care treatment at the same time), and share similar disease indications of the study treatment to avoid confounding variables. A tradeoff arises when defining the standard-of-care treatment. Active drugs as standard-of-care treatment (e.g., other active anti-asthma drugs as a control in the benchmark experiment) help us avoid confounding effects by disease indication (e.g., all patients in the cohort have asthma), but provide treatment effects of the study treatment that are marginal to the active drugs. No active drugs as standard-of-care treatment (e.g., no active anti-asthma drugs as control) allow us to estimate treatment effects by the significant contrast of the study treatment and control, but the confounding by indication remains [63,64].

## 6. Conclusion

Recently, many machine learning models for estimating HTE have been proposed, but there has been limited effort to apply the methods to a real-world healthcare problem. Our methodology review and benchmark study provided an overview of current methods and benchmark tests by applying them to the task of emulating clinical trials. We estimated the HTE of leukotriene receptor antagonists in reducing the risk of AD using nationwide healthcare claim data.

Our study has several limitations. First of all, our methodology comparison is only focused on the non-parametric approach using machine learning. Parametric HTE estimation methods we did not discuss in this review include generalized linear models with stratification-multilevel method or match-smoothing method [65]. Also, for our benchmark experiment, we could not set the time zero to separate the pre-treatment and post-treatment comorbidity because the first time the treatment was assigned is ambiguous in our short-term data. This might reduce treatment effect size. For future investigation, we plan to incorporate the interaction between treatment and post-treatment covariates in order to capture the underlying causal structure information among the variables and thus can lead to a more accurate and robust estimator of treatment effects [66].

Nonetheless, we delivered some insights that may benefit future research direction, such as that i) simple models work better if the true treatment effect is close to zero, which is the case in most clinical trials, ii) observational healthcare data, particularly administrative data (e.g., claim, EHRs) are incomplete, thus researchers need a caution for unobserved confounding variables, and iii) active drugs sharing the same indication to the treatment of interest as a placebo may help reduce confounding by indication. For future research direction, we envision that i) it is promising to develop HTE estimation methods that leverage the similar mechanism of treatment and placebo (e.g., active drugs sharing the same indication), either by the single learner or parameter sharing, ii) it is also critical to infuse human knowledge (e.g., PheWas,[55]) to complement the incomplete and sparse records in healthcare administrative data iii) As healthcare data always contains timestamps of those records, it is also worthwhile to incorporating temporal relationships of the variables when estimating HTE. Our benchmark experiment codes are publicly available to facilitate transparent comparison. As a future benchmark study, we will investigate the utility of HTE methods in clinical registries and trial data.

**FUNDING**

This work was supported by Cancer Prevention and Research institutes of Texas [RR180012 to X.J.]; the National Institute on Aging [R01AG066749]; the University of Texas system STARs Program; and the University of Texas Health Center startup.

## S.1 Supplementary method

### S.1.1 Reduce selection bias for average treatment effect estimation

The matching is to directly obtain counterfactual outcome by identifying propensity-matched neighbors; the stratification is to stratify subjects into groups with a similar level of propensity score and directly compare the outcomes within each group, and the inverse probability of treatment weighting (IPTW) is to down-weight over-sampled respondents and up-weight under-sampled respondents so that the treatment group and control group are similar.[1] Using the IPTW, the ATE for binary treatment T can be then estimated as

$$ATE = \frac{1}{N}\sum\left(\frac{TY}{e(X)} - \frac{(1-T)Y}{1-e(X)}\right), \qquad (1)$$

where $N$ is the number of total subjects and $e(X)$ is the propensity score at $T = 1$ given subject's feature $X$. The the treatment effect thus measures the amount of difference in outcome $Y$ due to intervention $T$ given the similarly weighted conditions $X$. We can obtain the propensity score $e(X)$ using arbitrary machine learning models (common choice is logistic regression or elastic net).

### S.1.2 Double machine learning

To briefly introduce the DML, we assume the outcome $Y$ and treatment assignment $T$ follow partially linear structural equations:

$$Y = \tau(X) \cdot T + g(X) + \epsilon, \text{ and}$$
$$T = f(X) + \eta, \qquad (2)$$

where $\epsilon$ and $\eta$ are noises that are independent to $X$ and $T$. The functions $g$ and $f$ are arbitrary supervised regression models. Our objective is to estimate the HTE $\tau(X)$. We can rewrite the structural equation as:

$$Y - E[Y|X] = \tau(X) \cdot (T - E[T|X]) + \epsilon, \qquad (3)$$

by applying expectation conditioned on $X$ and subtraction. In this equation, the unknowns are $E[Y|X]$ and $E[T|X]$, potential outcome and treatment assignment given the subject's feature $X$. Here we can estimate $E[Y|X]$ and $E[T|X]$ using any arbitrary supervised machine learning models, $f(X) = E[T|X]$ and $q(X) = E[Y|X]$**.** Once we obtain the two estimates, we denote $\hat{Y} = Y - q(X)$ and $\hat{T} = T - f(X)$. Now we can estimate the HTE via function $\hat{\tau}(X)$ using ordinary least square regression models:

$$\hat{\tau}(X) = argmin_{\tau} E[(\hat{Y} - \tau(X) \cdot \hat{T})^2], \qquad (4)$$

Note that the nuisance parameters $f$ and $g$ should be estimated in a separate set (i.e., cross fitting) to estimate. Theoretical error bounds are discussed in [2]. Sometimes researchers use other flexible machine learning models instead of linear regression when estimating $\hat{\tau}(X)$. These methods are called R (residual)-learners.

### S.1.3 Meta learner

S-learner fits a supervised model:

$$\mu(X, T) = E[Y|X, T], \qquad (5)$$

where the outcome estimation model can take any regression model (if the outcome is continuous) or classification models (if the outcome is a binary class). S-learner obtains the HTE by

$$\hat{\tau}(X) = \mu(X, T = 1) - \mu(X, T = 0), \qquad (6)$$

T-learner first trains "two" base learners to estimate the outcomes $Y$ of subjects $X$ under treatment $T$=1 and control *T=0*: $\mu_0(X) = E[Y(0)|X]$ and $\mu_1(X) = E[Y(1)|X]$, where the two outcome estimation models can take any supervised model. The second step is to calculate the difference between the estimated potential outcomes, which defines the HTE:

$$\hat{\tau}(X) = \hat{\mu}_0(X) - \hat{\mu}(t), \qquad (7)$$

X-learner takes three steps to estimate the HTE functions. First, it builds the outcome prediction models in treatment and control groups just like T-learner does:

$$\mu_0(X) = E[Y(0)|X], \mu_1(X) = E[Y(1)|X] \qquad (8)$$

Then, X-learner uses these two outcome prediction models to calculate estimated or pseudo HTE for subjects in the treatment group and control group respectively. Specifically, for subjects in the treatment group, we do not have *Y(0)*, the counterfactual outcome when the treated did not take the treatment. We use the outcome function $\mu_0(X)$ that is learned from the control group to make an estimation of the counterfactual outcomes. For subjects in the control group, we do not have $Y(1)$, the counterfactual outcome when the untreated ever took the treatment. We used the outcome function $\mu_1(X)$ that is learned from the treatment group to make an estimation of the counterfactual outcome. Then the "pseudo" HTE is defined as: $\hat{\tau}_1 = Y(1) - \mu_0(X)$ for the treated $(T = 1)$ and $\hat{\tau}_0 =$ $\mu_1(X) - Y(0)$ for the untreated ($T = 0$). This is analogous to missing value imputation in machine learning. The counterfactual outcomes are always unobserved (by definition). We impute the missing counterfactual outcome with the estimator learned from subjects having the outcomes. Now we fit separate supervised models for HTE:$\hat{\tau}_1(X)$ with input $X$ and output $\hat{\tau}_1$ for the treated ($T = 1$) and $\hat{\tau}_0(X)$ with input $X$ and output $\hat{\tau}_0$ for the untreated ($T = 0$). In other words, $\hat{\tau}_1(X)$ is a function to estimate the HTE when the subjects in the training data are all treated; $\hat{\tau}_0(X)$ is a function to estimate the HTE when the subjects in the training data are all not treated. The final step is to mix the two HTE functions to obtain the final HTE function:

$$\hat{\tau}(X) = w(X) \cdot \hat{\tau}_0(X) + (1 - w(X)) \cdot \hat{\tau}_1(X), w(X) \in [0,1], \quad (9)$$

which is a weighted sum with a weighting function $w(x)$. Here controls the contribution of HTE derived from different situations: either when the subjects are all treated or untreated. If $w(x)$ is the probability of being treated (a.k.a. Propensity score), will be high if there are many treated subjects and fewer untreated subjects (or vice versa). So the HTE function derived from a large group will contribute less to $\hat{\tau}(X)$; the HTE function derived from a small group will contribute more to $\hat{\tau}(X)$.

**S.1.4 IF-PEHE**

The precision of estimating heterogeneous effects (PEHE) [3] is defined as:

$$l(\hat{T}) = \left|\left|\hat{\tau}(X) - \tau(X)\right|\right|_{\theta}^{2}, \quad (10)$$

where $\hat{\tau}(X)$ represents the estimated HTE function and $\tau(X)$represents the true HTE function. However, we never observe the $\tau(X)$ in real-world observational data and thus, we cannot calculate PEHE directly. Ahmed et al.[4] proposed a method that uses a Tylor-like expansion to approximate the true PEHE.

Their validation procedure can be split into two steps: the first step is to build a plug-in estimation model; and then, use the von Mises expansion to make compensation for plug-in bias. To simplify notation, we call it IF-PEHE. Only first-order influence functions are considered to be used to approximate the loss function which has the form

$$l_{\theta}(\hat{\tau}) \approx l_{\underline{\theta}}(\hat{\tau}) + E_{\theta}[l_{\underline{\theta}}^{(1)}(z; \hat{\tau})], \quad (11)$$

where $l_{\underline{\theta}}^{(1)}(\hat{T})$denotes the first-order derivatives of the loss function regarding the plug-in estimate of the HTE which is also known as the first-order influence function. IF-PEHE provided a specific method to calculate the first-order influence function. Given a dataset, $Z = (Y, T, X)$, where Y is the response variable, T is a binary variable indicating treatment, and X is the matrix of features. $e(x) = p(W|X = x)$is a propensity score function. Then the first-order influence function can be calculated by

$$l_{\underline{\theta}}^{(1)}(z; \hat{\tau}) = (1 - 2T(T - e(x)) \cdot C^{-1})\underline{\tau}^2(x) + 2T(T - e(x)) \cdot C^{-1}Y - A(\underline{\tau}(x) - \hat{\tau}(x))^2 + \hat{\tau}^2(x). \quad (12)$$

**S.2 Supplementary: Benchmark experiment details**

**S.2.1 Treatment definition**

**Table S1.** Treatment of interest. Five types of anti-asthma drugs. We set leukotriene receptor antagonist as the study treatment and set the remaining anti-asthma drugs (Beta-2 adrenergic receptor agonist, Anticholinergic drug, Xanthine, and Corticosteroid) as standard-of-care treatment.

| Class | Drugs (Drugbank ID) |
|---:|---:|
| Beta-2 adrenergic receptor agonist | Salbutamol (DB01001), Levosalbutamol (DB13139), Vilanterol (DB09082), Arformoterol (DB01274), Carmoterol (DB15784), Indacaterol (DB05039), Orciprenaline (DB00816), Reproterol (DB12846), Salmeterol (DB00938), Terbutaline (DB00871), Formoterol (DB00983) |
| Anticholinergic drug | Umeclidinium (DB09076), Ipratropium (DB00332) |
| Xanthine | Theophylline (DB00277), Oxtriphylline (DB01303), Aminophylline (DB01223) |
| Corticosteroid | Fluticasone (DB13867), Budesonide (DB01222), Mometasone |
| | (DB00764), Beclomethasone dipropionate (DB00394), Flunisolide (DB00180), Ciclesonide (DB01410), |
| Leukotriene receptor antagonist | Montelukast (DB00471), Zileuton (DB00744), Zafirlukast (DB00549), Pranlukast (DB01411) |

**S.2.2 Outcome**

The outcome of interest was ADRD onset, which we detected as having either an ADRD diagnosis code or medication. The ADRD diagnosis codes were PheWas codes for 290.11 (Alzheimer's disease); 290.12 (Frontotemporal dementia, Pick's disease, Senile degeneration of brain); 290.13 (Senile dementia); and 290.16 (Vascular dementia, Vascular dementia with delirium/delusions/depressed mood). The ADRD medications were acetylcholinesterase inhibitors (Donepezil, Galantamine, and Rivastigmine) or memantine. The definition of ADRD in EHRs can be controversial considering the fact that EHRs are for billing purposes.

**S.2.3 Evaluation measure**

To calculate the IF-PEHE on the dataset, we follow the same experiment setting as Ahmed et al. [4] did in their experiments:

- Step 1: Train 2 XGBoost Classifiers to estimate $\mu_1, \mu_0$. And then calculate the plug-in estimate for by $\underline{\tau} = \mu_1 - \mu_0$; use the random forest algorithm to estimate the propensity score function $\underline{\pi}(x)$
- Step 2: Calculate the influence function $l_{\underline{\theta}}^{(1)}(Z; \hat{\tau})$ and then we can get IF-PEHE by $\hat{l} = \sum_{i \in Z_p} [(\hat{\tau}(X_i) - \underline{\tau}(X_i))^2 + l_{\underline{\theta}}^{(1)}(Z_i; \hat{\tau})]$

**S.2.4 Propensity score estimation model**

Propensity score measures the likelihood of a patient to take a treatment and arbitrary supervised models can be utilized to estimate propensity scores. We tried four different machine learning algorithms: logistic regression, random forest, gradient boosting classifier, and XGBoost. We evaluated the propensity scores by checking the models' AUROC scores (Table S2). Low AUROC scores indicate either that the model does not learn much about the treatment assignment mechanism (underfitting) or that predicting treatment assignment is difficult (treatment randomly assigned within the data). If a model's AUROC is larger, it indicates the model can tell the treatment assignment bias

Table S2. Comparison of different models' performance on estimating propensity score P(T=1|X).

| Models | AUROC |
|---|---|
| Logistic regression | 0.5472 |
| Random Forest | **0.5774** |
| Gradient Boosting Classifier | 0.5769 |
| XGBoost Classifier | 0.5736 |

**S.2.5 Feature selection for HTE modeling**

We performed filter methods with three kinds of divergence measurements on our best model, random forest based S-learner, to check if feature selection can help improve the performance of our model. We found that with regard to S-learner, the performance reached its maximum when all the features were included in to train the model (Fig. S1).

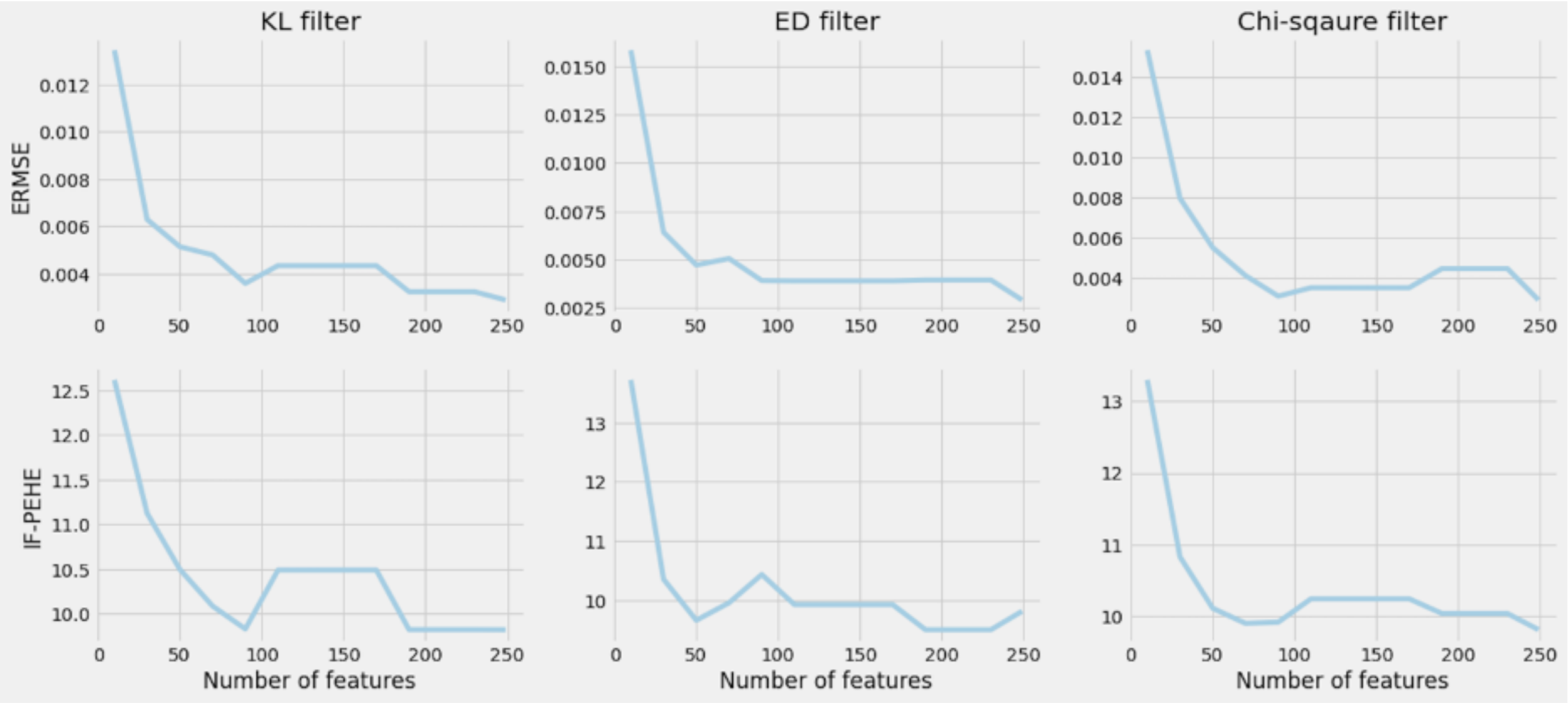

**Figure S1.** The performance of S-learner with random forest base learner given selected features. The features were selected using three kinds of filter methods. We measured the HTE estimation performance using ERMSE and IF-PEHE.